\documentclass[letterpaper]{article} 
\usepackage{aaai25}  
\usepackage{times}  
\usepackage{helvet}  
\usepackage{courier}  
\usepackage[hyphens]{url}  
\usepackage{graphicx} 
\usepackage{multirow}
\usepackage{color}
\usepackage{amsmath}
\usepackage{amssymb}
\usepackage{booktabs}
\usepackage{natbib}  
\usepackage{caption} 
\usepackage[ruled,vlined]{algorithm2e}
\definecolor{commentcolor}{RGB}{110,154,155}   
\newcommand{\PyComment}[1]{\ttfamily\textcolor{commentcolor}{\# #1}}  
\newcommand{\PyCode}[1]{\ttfamily\textcolor{black}{#1}} 

\usepackage{newfloat}
\usepackage{listings}
\DeclareCaptionStyle{ruled}{labelfont=normalfont,labelsep=colon,strut=off} 
\title{Out of Length Text Recognition with Sub-String Matching}
\author {
    Yongkun Du\textsuperscript{\rm 1},
    Zhineng Chen\textsuperscript{\rm 1}\thanks{Corresponding Author},
    Caiyan Jia\textsuperscript{\rm 2},
    Xieping Gao\textsuperscript{\rm 3},
    Yu-Gang Jiang\textsuperscript{\rm 1}
}
\affiliations {
    \textsuperscript{\rm 1}School of Computer Science, Fudan University, China\\
    \textsuperscript{\rm 2}School of Computer Science and Technology, Beijing Jiaotong University, China\\
    \textsuperscript{\rm 3}Laboratory for Artificial Intelligence and International Communication, Hunan Normal University, China\\
ykdu23@m.fudan.edu.cn, \{zhinchen, ygj\}@fudan.edu.cn, cyjia@bjtu.edu.cn, xpgao@hunnu.edu.cn
}

\usepackage{bibentry}

\begin{document}

\maketitle

\begin{abstract}
Scene Text Recognition (STR) methods have demonstrated robust performance in word-level text recognition. However, in real applications the text image is sometimes long due to detected with multiple horizontal words. It triggers the requirement to build long text recognition models from readily available short (i.e., word-level) text datasets, which has been less studied previously. In this paper, we term this task Out of Length (OOL) text recognition. We establish the first Long Text Benchmark (LTB) to facilitate the assessment of different methods in long text recognition. Meanwhile, we propose a novel method called OOL Text Recognition with sub-String Matching (SMTR). SMTR comprises two cross-attention-based modules: one encodes a sub-string containing multiple characters into next and previous queries, and the other employs the queries to attend to the image features, matching the sub-string and simultaneously recognizing its next and previous character. SMTR can recognize text of arbitrary length by iterating the process above. To avoid being trapped in recognizing highly similar sub-strings, we introduce a regularization training to compel SMTR to effectively discover subtle differences between similar sub-strings for precise matching. In addition, we propose an inference augmentation strategy to alleviate confusion caused by identical sub-strings in the same text and improve the overall recognition efficiency. Extensive experimental results reveal that SMTR, even when trained exclusively on short text, outperforms existing methods in public short text benchmarks and exhibits a clear advantage on LTB.

\end{abstract}
%
\begin{links}
    \link{Code}{https://github.com/Topdu/OpenOCR}
\end{links}

\section{Introduction}

\begin{figure}[t]
  \centering
\includegraphics[width=0.45\textwidth]{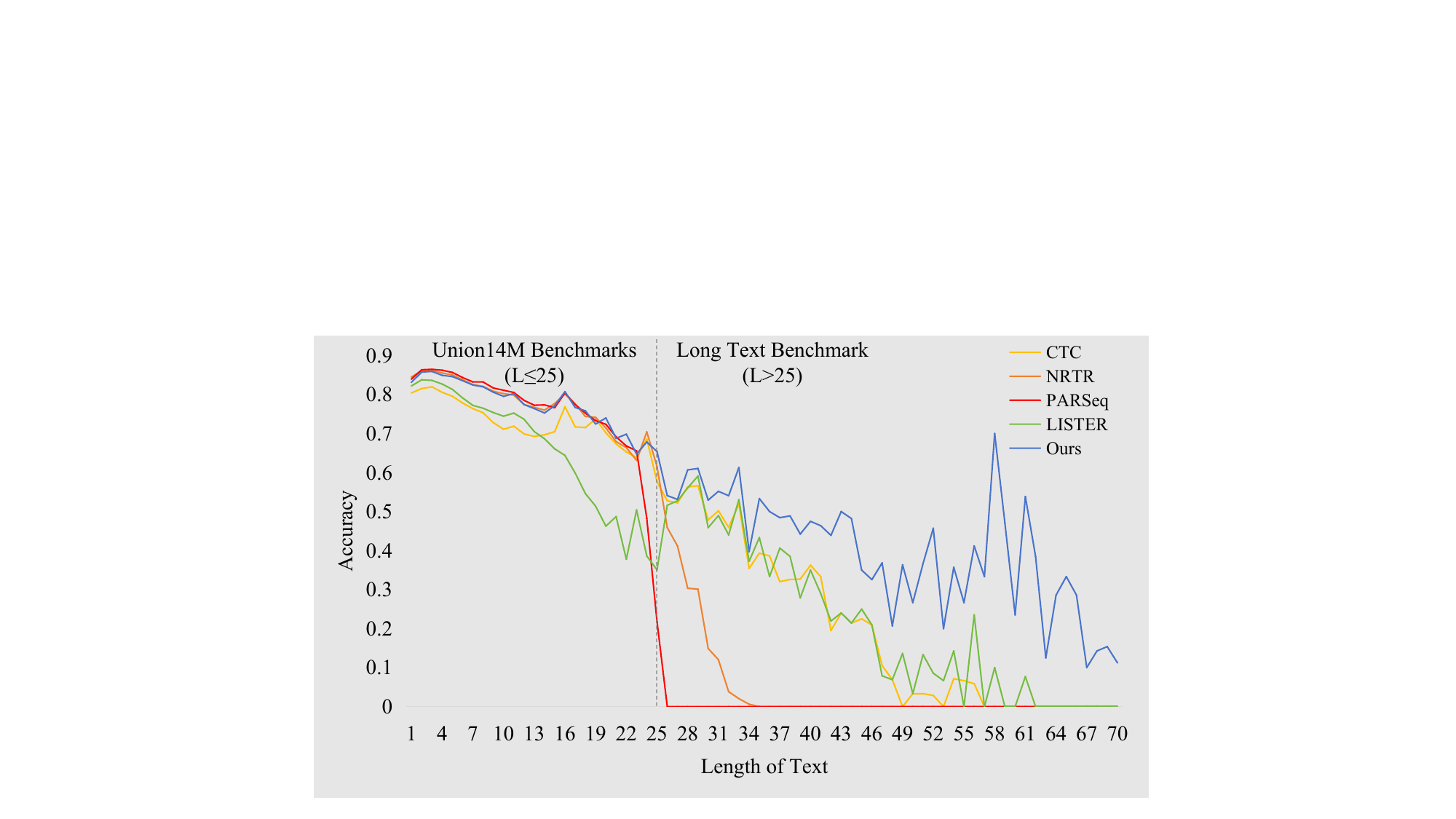} 
\caption{Attention-based methods like NRTR \cite{Sheng2019nrtr} and PARSeq \cite{BautistaA22PARSeq} perform well in short text recognition, while the CTC-based FocalSVTR (CTC) and the length-insensitive LISTER \cite{iccv2023lister} outperform them in long text but worse in short text. Our proposed SMTR gets superior performance in both short and long text recognition.}
\label{fig:fig1}
\end{figure}

Extracting text from natural images is a crucial and well-established task, typically encompassing scene text detection and recognition. In scene text recognition (STR), past research has achieved significant progress in word-level text recognition. However, in real-world applications, the text is not always detected as individual words, but a line-level text sometimes. That is, the text instance detected by scene text detectors may contain multiple horizontal words, presenting a long text recognition challenge. It is worth noting that existing STR datasets are predominantly compromised of word-level text, where long text like the aforementioned is scarce. Therefore, accurately recognizing long text by utilizing solely short (i.e., word-level) datasets has emerged as a promising yet challenging frontier in STR, which we term Out of Length (OOL) text recognition in this paper.

Existing STR models are mostly designed towards recognizing short text with no more than 25 characters, i.e., $L \leq 25$. We establish the first Long Text Benchmark (LTB) that focuses on the long text ($L >$ 25). As shown in Fig. \ref{fig:fig1}, we evaluate several popular STR models \cite{Sheng2019nrtr,duijcai2022svtr,BautistaA22PARSeq,iccv2023lister} on LTB. The results show that the CTC-based method \cite{duijcai2022svtr} and LISTER \cite{iccv2023lister} can recognize long text due to their length extrapolation capabilities. However, their accuracy declines rapidly as text length increases. Additionally, these methods show a noticeable performance gap in short text recognition compared to the competitors \cite{Sheng2019nrtr,BautistaA22PARSeq}, due to not being equipped with an advanced decoder. On the other hand, NRTR \cite{Sheng2019nrtr} and PARSeq \cite{BautistaA22PARSeq}, as representatives of the attention-based methods, employ the self-attention mechanism to encode the decoded characters along with absolute positional information \cite{NIPS2017_attn}. It serves as the linguistic context or position-involved representation to aid decoders in accurately recognizing characters. This powerful approach enables these methods to achieve state-of-the-art results in short text recognition. However, they are unable to effectively learn the representation beyond the length of training text, which limits them from processing long text when trained solely on short text. Similar limitations are also observed in other attention-based methods \cite{shi2019aster,Wang_2021_visionlan,fang2021abinet,du2023cppd,ijcai2023LPV,zheng2024cdistnet}.

In this paper, we introduce a novel method termed OOL Text Recognition with sub-String Matching (SMTR), which achieves OOL text recognition by innovatively leveraging string-matching techniques. Specifically, SMTR recognizes text by first matching a specified sub-string within an image, and then positioning and recognizing the Next and Previous characters of the sub-string.  SMTR can recognize text of arbitrary length by iterating the process above. It gets rid of the absolute position and recognizes text fully relying on sub-string identification and extrapolation. This new paradigm is reasonable as both short and long text can be decomposed into a series of sub-string units. The matching process only compares a portion of the text image with the sub-string, regardless of whether the whole text is short or long. Based on this fact, SMTR learns to match sub-strings by using only short text images but can still recognize long text.

To implement this, SMTR develops a lightweight architecture with a sub-string encoder and a sub-string matcher. The former encodes a sub-string as two representations referred to as next and previous queries. While the latter directs the two queries to attend to visual features, aiming to accurately match the sub-string and recognize the next and previous characters. In this new recognition paradigm, similar or repeated sub-strings in one text image inevitably hinder precise sub-string matching. To effectively screen for similar sub-strings, we propose a regularization training strategy that encourages SMTR to pay attention to the subtle differences between sub-strings, thus better distinguishing them. In addition, we introduce an inference augmentation strategy which breaks a long text image into multiple sub-images for independent processing, thus significantly alleviating the side affection caused by repeated sub-strings. Moreover, by processing these sub-images in parallel, SMTR improves the overall recognition efficiency remarkably. Experimental results demonstrate that SMTR achieves highly competitive performance on challenging short text benchmarks and exhibits a noteworthy advantage on LTB. The contributions are summarised as follows:

\begin{itemize}
\item We, for the first time, term the requirement of building long text recognition models based on short text datasets, as OOL text recognition challenge, and establish a long text benchmark called LTB to assess the long text recognition capability of STR models.

\item We propose SMTR, a novel method that elegantly incorporates string-matching techniques to address the OOL challenge. Meanwhile, regularization training and inference augmentation strategies are proposed to ensure precise recognition following this pipeline.

\item SMTR achieves state-of-the-art performance on both public short text benchmark and LTB, utilizing merely short text for training. Our study enriches the STR schemes in handling diverse real-world applications. 
\end{itemize}

\section{Related Work}

Existing STR methods mainly focus on word-level text recognition. Among them, attention-based \cite{lee2016attention} methods are intensively studied for their impressive performance. Some of these methods \cite{BautistaA22PARSeq,yue2020robustscanner,Wang_2021_visionlan,du2023cppd,fang2021abinet,ijcai2023LPV,TPAMI2022ABINetPP,zhao24dptr,du2024igtr} use learnable position embeddings to accurately learn context information. However, because learnable position embeddings are typically not scalable, these methods can only recognize short text. In contract, methods \cite{shi2019aster,li2019sar,Sheng2019nrtr,qiao2021pimnet,zheng2024cdistnet,tip2024iast} employ LSTM \cite{lstm,gru} or masked self-attention mechanism with sinusoidal positional encoding \cite{lee2016attention} to model context. They have some length extrapolation capability. Nevertheless, these models are primarily trained on short text. Therefore their inference ability is limited when dealing with long text, resulting in a significant decrease in recognition accuracy. Additionally, some methods do not rely on well-designed attention-based decoders, such as CTC-based \cite{CTC} methods \cite{shi2017crnn,hu2020gtc,duijcai2022svtr} and length-insensitive LISTER \cite{iccv2023lister}. They do not perform well on short text recognition although exhibiting slightly stronger capability in long text recognition.

SMTR circumvents the above problems and recognizes text by sub-string matching. This scheme ensures SMTR's applicability to long text even when only seen short text images, as sub-string is an essential component for both short and long text.

\section{Method}

\begin{figure*}
  \centering
\includegraphics[width=0.94\textwidth]{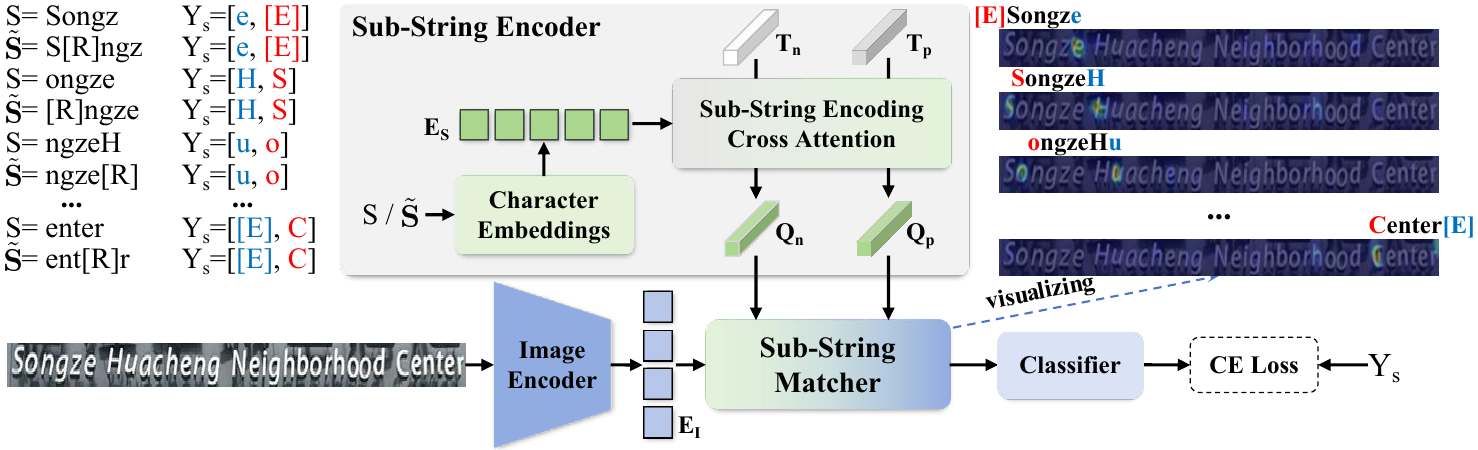} 
  \caption{Overview of SMTR. SMTR is a lightweight recognizer consisting of two cross-attention-based modules, i.e., sub-string encoder and sub-string matcher. $S$ and $\tilde{S}$ denote a sub-string and it regularized counterpart. $Y_s$ denotes the next-previous character set of the sub-string. [R] denotes a randomly selected character from the character set.}
  \label{fig:OOL}
\end{figure*}

Fig. \ref{fig:OOL} illustrates the overall architecture of SMTR. Given a text image $X \in \mathbb{R}^{3 \times H \times W}$, the image encoder extracts its image embeddings $E_I$. A sub-string (S) is encoded into both next and previous hidden representation ($Q_n$ and $Q_p$) by the sub-string encoder. $Q_n$ and $Q_p$ are then fed into the sub-string matcher, which implicitly matches the sub-string in the text image and recognizes its next and previous characters bidirectionally. In addition, we discuss issues raised within this new string-matching-based paradigm and propose our solutions.

\subsection{Image Encoder}

To accommodate various aspect ratios of the input $X$, we develop a dedicated image encoder termed FocalSVTR to extract visual features as in LISTER \cite{iccv2023lister}. Firstly, the input $X$ is divided into patch embeddings ($\in \mathbb{R}^{C_0 \times \frac{H}{4} \times \frac{W}{4}}$) by two convolution with stride 2. Then, following SVTR \cite{duijcai2022svtr}, three stages comprising [6, 6, 6] layers of focal modulation \cite{YangLDG22focalnet} are stacked. At the end of the first and second stages, $C_0$ is extended with a factor of 2 by a convolution. In particular, at the end of the second stage, the convolution downsamples the height of the patch embeddings to $\frac{H}{8}$. Finally, the output features ($\in \mathbb{R}^{C \times \frac{H}{8} \times \frac{W}{4}}$) are flattened to obtain image embeddings $E_I$ $\in \mathbb{R}^{C \times \frac{HW}{32}}$, where $C=4C_0$.

\subsection{Sub-String Encoder}

\noindent\textbf{Sub-String and Label Generation}. For a labled text image, all sub-strings $S$ and their next-previous character sets $Y_s = [Y_n, Y_p]$ can be generated as follows. Taking ``datours'' in Fig. \ref{fig:long_inference} as an example. It is first prefixed and suffixed with ``[B]$_{l_s}$'', resulting in the formatted label: ``[B]$_{l_s}$datours[B]$_{l_s}$''. Here, ``[B]$_{l_s}$'' represents a string with \(l_s\) blank characters [B], and \(l_s\) is the predefined sub-string length. After that, the formatted label is traversed by a window of size $l_s$ to obtain all sub-strings and their \(Y_s\) labels. If a sub-string contains [B] on one side, the corresponding label in \(Y_s\) is replaced with [P], indicating that there is no additional character on that side.

Fig. \ref{fig:long_inference} shows all sub-strings of ``datours'' and their corresponding $Y_s$ when $l_s$ is set to 5. The sub-string is initialized with ``[B]$_5$'' and \(Y_s\)=[d, s], indicating the start of the next and previous inferences. For a text of length \(L\), the number of sub-strings is $2L+1$ when $L < l_s$, and $L + l_s$ otherwise.

\noindent\textbf{Sub-String Encoding}. Given a sub-string $S$, firstly, each character in $S$ is converted into a \emph{C}-dim embedding via a character embedding layer, and the sub-string embedding $E_s \in \mathbb{R}^{l_s \times C}$ is obtained by feature concatenation. Then, the next token $T_n \in \mathbb{R}^{1 \times C}$ is introduced, which is a shared and learnable token. $T_n$ acts as a query and performs cross-attention with $E_s$, obtaining the next sub-string hidden representation $Q_n \in \mathbb{R}^{1 \times C}$. Similarity, the previous token $T_p\in \mathbb{R}^{1 \times C}$ is introduced and generates the previous sub-string hidden representation $Q_p\in \mathbb{R}^{1 \times C}$. Assuming there are $h$ attention heads, each with different parameters $W_i^q, W_i^k, W_i^v \in \mathbb{R}^{C \times C_h}$, $i$ denotes the index of the attention head. The above process can be formulated as:
\begin{gather}
    Q_{z}= \text{MHead}(T_{z}, E_s, E_s) + T_{z} \notag \\
    \text{MHead}(Q, K, V) = \text{Concat}(\text{head}_1, \text{head}_2, \ldots, \text{head}_h)  \notag \\
\text{head}_i = A_i(VW_i^v) \notag \\ 
A_i = \sigma \left((QW_i^q)\left(KW_i^k\right)^tC_h^{-0.5}\right) \in \mathbb{R}^{1 \times \frac{HW}{32}}
\end{gather}
\noindent where $z$ is $n$ or $p$, $C_h$ = $\frac{C}{h}$, and $\sigma$ is the Softmax function.

\subsection{Sub-String Matcher}

We use one cross-attention \cite{NIPS2017_attn} to implement sub-string matching, where the sub-string hidden representation $Q_n$ or $Q_p$ as query and image embeddings $E_I$ as key and value.  The process can be expressed as:
\begin{equation}
\text{Matcher}(Q_z, E_I) = \text{MHead}(Q_z, E_I, E_I)
\end{equation}

Taking $Q_n$ as an example, the sub-string matcher computes the attention map $A_i$, which focuses on the position of the next character of the sub-string, as shown in the attention maps in Fig. \ref{fig:OOL}. Consequently, the output of the sub-string matcher represents the next character feature $F_n = \text{Matcher}(Q_{n}, E_I) \in \mathbb{R}^{1 \times C}$, which then undergoes a Classifier ($ \in \mathbb{R}^{C \times (N_c +1)}$) to generate prediction $\tilde{Y}_n \in \mathbb{R}^{1 \times (N_c +1)}$, i.e., $\tilde{Y}_n = \text{Classifier}(F_n)$, where ${+1}$  is for the end symbol [E]. In the same way, the previous character prediction is obtained by $\tilde{Y}_p = \text{Classifier}(F_p), F_p=\text{Matcher}(Q_{p}, E_I)$.

\subsection{Optimization Objective}

During training, for a text instance with \emph{N} sub-strings, SMTR predicts the next and previous characters ($\tilde{Y}_n$ and $\tilde{Y}_p$) for each sub-string. The loss $\mathcal{L}$ is computed by comparing the label $Y_s = [Y_n, Y_p]$ with $\tilde{Y}_s = [\tilde{Y}_n, \tilde{Y}_p]$:
\begin{align}
\mathcal{L} = \frac{1}{K} \sum_{i=1}^{N} (ce(\sigma(\tilde{Y}_n^i), Y_n^i) + ce(\sigma(\tilde{Y}_p^i), Y_p^i))\\
ce(\tilde{y}, y) = \left\{  
             \begin{array}{lr}
            -\sum_{c=1}^{N_c}y_{c}\log(\tilde{y}_{c}), ~ &y \neq \text{[P]}.\\  \\
             0, &y = \text{[P]}.     
             \end{array}  
\right. \notag
\end{align}  
\noindent where label [P] is not involved in the loss computation, $K$ is the number of valid loss terms. It is equal to $2N-2L$ when $L < l_s$, and $2N-2(l_s-1)$ otherwise.

\subsection{Regularization Training}

SMTR performs matching and recognition by aligning the sub-string embedding within the image feature space. However, this process can be compromised by similar or repeated sub-strings, as they provide quite similar visual features and context. We categorize this problem as similar distraction and repeated confusion, and propose two strategies to alleviate them. The first is regularization training for similar distraction described as follows.

\begin{figure}[t]
  \centering
\includegraphics[width=0.44\textwidth]{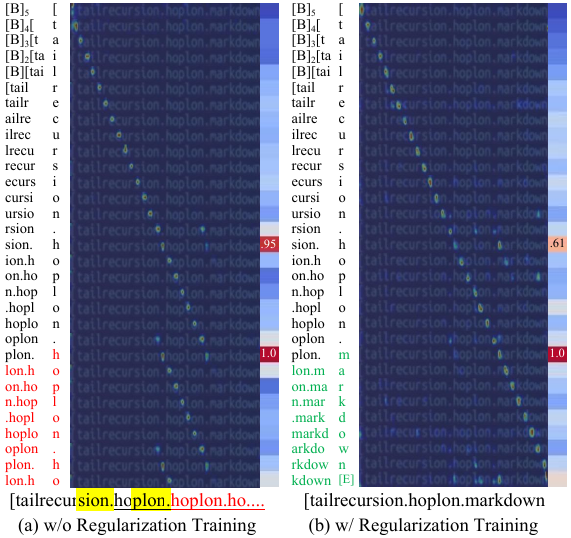} 
\caption{Attention maps of SMTR w/o (left) and w/ (right) the proposed regularization training, which rectifies recognition errors (red) caused by similar sub-strings (yellow).}
\label{fig:regular_training}
\end{figure}

SMTR is easily disturbed by sub-strings with similar visual appearance and generates error recognition. As shown in Fig. \ref{fig:regular_training}(a), the next character after ``plon.'' is mistakenly identified as the next character of ``sion.'', resulting in an undesired circular recognition. To overcome this issue, SMTR should be endowed with the ability to discover subtle differences between similar sub-strings. To this end, we propose regularization training (RT), which alleviates this problem by generating similar sub-strings $\tilde{S}$ of $S$ for training reinforcement. As depicted in the top-left side of Fig. \ref{fig:OOL}, one character in $S$ is replaced with [R], a character randomly selected from the character set, to obtain $\tilde{S}$, and keep its label $Y_s$ unchanged. By doing this, the number of similar sub-strings is largely enriched during training. SMTR is compelled to leverage all the $l_s$ characters, rather than a few nearby ones, to distinguish similar sub-strings appearing in the same text image, which therefore can be better identified.

\begin{figure}
  \centering
\includegraphics[width=0.46\textwidth]{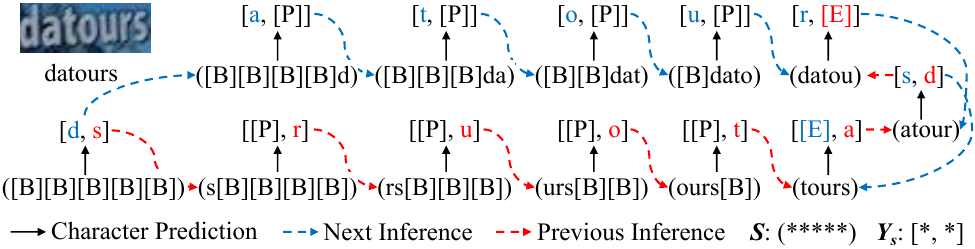} 
  \caption{Illustration of SMTR base inference process, where [E] means the end token for inference termination.}
  \label{fig:long_inference}
\end{figure}

\begin{algorithm}[t] \footnotesize
\caption{Base Inference Pseudo-code in Python}
\PyCode{def Inference(Img,Tok,Ss=[0]*$l_s$,Se=[]):} \\
\Indp
    \PyComment{Img: Input Image, $3 \times H \times W$.} \\
    \PyComment{Tok: Next or Previous Token, $1 \times C$.} \\
    \PyComment{Ss/Se: The start/end sub-string.} \\
    \PyComment{[0]: The [B].} \\
    \PyComment{$l_s$: The length of sub-string.} \\
    \PyCode{Result = [], S = Ss} \\
    \PyCode{IE = ImageEncoder(Img) \PyComment{$\frac{HW}{32} \times C$}}  \\
    \PyCode{While True:} \\
    \Indp
        \PyCode{Q = SubStringEncoder(Tok, S)} \\
        \PyCode{O = SubStringMatcher(Q,IE)} \\ 
        \PyCode{Char = Classifier(O).argmax(-1)} \\
        \PyCode{if Tok is $T_n$:} \\
        \Indp
            \PyCode{S = S[1:] + [Char]} \\
        \Indm
        \PyCode{else:} \\
        \Indp
            \PyCode{S = [Char] + S[:1]} \\
        \Indm
        \PyComment{EOS: The index of end symbol.} \\
        \PyCode{if Char == EOS or S == Se:} \\
        \Indp
            \PyCode{break} \\
        \Indm
        \PyCode{Result.append(Char)} \\
    \Indm
    \PyCode{return Result} \\
    \PyComment{Result is the recognition result.} \\
\Indm
\label{alg:inference}
\end{algorithm}

\begin{table}[t]\footnotesize
\centering
\setlength{\tabcolsep}{5pt}{
\begin{tabular}{l|cc|c}
\hline
        Dataset           & Samples  & Repeat $S$ &        \\
\hline
Union14M-L Training  & 3,230,742 & 885           & 0.03\% \\
Union14M-Benchmark & 409,383 & 22            & 0.01\% \\
Common Benchmarks   & 7,248    & 0             & 0.00\% \\
LTB w/o IA                & 4,789    & 472           & 9.86\% \\
LTB w/ IA        & 4,789    & 20            & 0.42\% \\
\hline
\end{tabular}}
\caption{The percentage of text instances with repeated sub-strings ($l_s\geq 5$) in different datasets. Inference augmentation (IA) drastically reduces the percentage in LTB.}
\label{tab:repeatss}
\end{table}

\subsection{Inference Augmentation}

We then introduce the inference augmentation (IA) proposed to mitigate the repeated confusion as follows.

We first introduce the base inference of SMTR. Since SMTR does not know which sub-strings are contained in the text image, it implements the decoding inference with the sub-strings beginning with [B]$_{l_s}$, and iteratively predicts the next and previous characters as shown in Fig. \ref{fig:long_inference}. The process is terminated when the end token [E] is predicted. Alg. \ref{alg:inference} provides a detailed description of this process.

\begin{figure}
  \centering
\includegraphics[width=0.47\textwidth]{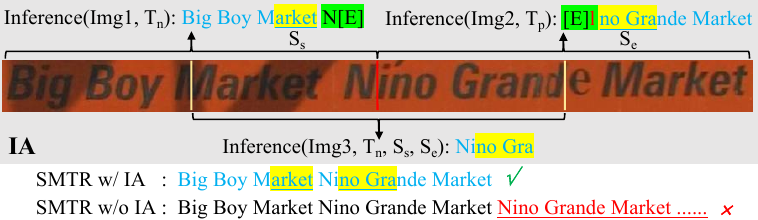} 
  \caption{Illustration of inference augmentation (IA) for long text recognition. Text spaces are manually inserted.}
  \label{fig:ultra_long_inference}
\end{figure}

The base inference fails to distinguish repeated sub-strings in one image, such as the \emph{SMTR w/o IA} result in Fig. \ref{fig:ultra_long_inference}. Tab. \ref{tab:repeatss} shows the statistics of repeated sub-strings in typical short text datasets and LTB. The high percentage of repeated sub-strings (near 10\%) in LTB seriously affects the recognition. Consequently, we propose IA that first slices the long text image into three short sub-images and then employs a split-merge scheme for recognition. This process is depicted in Fig. \ref{fig:ultra_long_inference}. Specifically, the text image is sliced into three sub-images whose width is halved, i.e., the left sub-image \texttt{Img1}, the right sub-image \texttt{Img2} and the central sub-image \texttt{Img3}. Note that both \texttt{Img1} and \texttt{Img2} are overlapped with \texttt{Img3}. Then, by taking advantage of the bidirectional recognition property of SMTR, we recognize \texttt{Img1} and \texttt{Img2} following the next and previous prediction paths, respectively, i.e., \texttt{Inference(Img1, $T_n$)} and \texttt{Inference(Img2, $T_p$)}. The process obtains the results shown in the top line of Fig. \ref{fig:ultra_long_inference} in roughly half of the inference iterations, as both predictions can be carried out in parallel. Since the slice may operate on characters and lead to mis-recognition, we pick out recognized sub-strings also appearing in \texttt{Img3} from both sides according to attention maps (highlighted by yellow in Fig. \ref{fig:ultra_long_inference}), i.e., $S_s$ and $S_e$. In the following, we predict \texttt{Img3} with \texttt{Inference(Img3, $T_n$, $S_s$, $S_e$)}, which gets the central characters. Finally, the full result, e.g., the \emph{SMTR w/ IA} result in Fig. \ref{fig:ultra_long_inference}, is obtained by a simple post-processing.

It is seen in Tab. \ref{tab:repeatss} that IA drastically reduces the percentage of repeated sub-strings in LTB from nearly 10\% to 0.4\%, thus significantly alleviating the repeated confusion. The effectiveness of IA will be demonstrated in Tab. \ref{tab:ltb_result}.

\subsection{Long Text Benchmark}

Existing public datasets \cite{jung2011touchkaist,nagy2012neocr,veit2016cocotext,shi2017icdar2017rctw,zhang2017ubertext,yuliang2017detectingctw1500,mathew2017benchmarkingiiitilst,he2018icpr2018mtwi,nayef2019icdar2019mlt19,sun2019icdarlsvt,chng2019icdar2019art,zhang2019icdarrects,krylov2021openintelocr,singh2021textocr,long2022towardshiertext} predominantly consist of short text ($L\leq$ 25). However, they also have a few instances with more than 25 characters. These instances are usually regarded as noise and discarded in developing traditional short text models. We collect these instances and construct a Long Text Benchmark (\textit{LTB}). \textit{LTB} contains nearly 4.8k samples whose length is above 25. Since all samples are excluded from the training process, \textit{LTB} establishes a new benchmark exclusively for evaluating the performance of STR models on long text recognition.
To assess the impact of length variation on recognition, we divide \textit{LTB} into three parts based on text length: [26, 35], [36, 55], and $\geq$ 56, as shown in Tab. \ref{tab:ltb_result}. 

\section{Experiments}

\subsection{Datasets and Implementation Details}
\label{sec:Implementation}

We evaluate SMTR on both English and Chinese datasets. For English, our models are trained on Union14M-L \cite{jiang2023revisiting}, which contains about 3.2 million real-world text images ($L \leq 25$). Then, the models are tested on both LTB and two short text benchmarks: (1) Common benchmarks, i.e., ICDAR 2013 (\textit{IC13}) \cite{icdar2013}, Street View Text (\textit{SVT}) \cite{Wang2011SVT}, IIIT5K-Words (\textit{IIIT}) \cite{IIIT5K}, ICDAR 2015 (\textit{IC15}) \cite{icdar2015}, Street View Text-Perspective (\textit{SVTP}) \cite{SVTP} and CUTE80 (\textit{CUTE}) \cite{Risnumawan2014cute}. For \textit{IC13} and \textit{IC15}, we use the versions with 857 and 1,811 images, respectively. (2) Union14M-Benchmark, which includes seven challenging subsets: Curve (\textit{CUR}), Multi-Oriented (\textit{MLO}), Artistic (\textit{ART}), Contextless (\textit{CTL}), Salient (\textit{SAL}), Multi-Words (\textit{MLW}) and General (\textit{GEN}). Note that the 6,600 samples overlapping between Union14M-L and Union14M-Benchmark are filtered \cite{du2024svtrv2}. For Chinese, we use Chinese text recognition (CTR) dataset \cite{chen2021benchmarking}, which contains four subsets: \textit{Scene}, \textit{Web}, \textit{Document} (\textit{Doc}) and \textit{Hand-Writing} (\textit{HW}). We train the model on the whole training set and use \textit{Scene} validation subset to determine the best model, which is assessed on the test subsets.

We use AdamW optimizer \cite{adamw} with a weight decay of 0.05 for training. The LR is set to $6.5\times 10^{-4}$ and batchsize is set to 1024. One cycle LR scheduler \cite{cosine} with 1.5/4.5 epochs linear warm-up is used in all the 20/100 epochs, where a/b means a for English and b for Chinese. Regarding the aspect ratio, all images are resized to a maximum pixel size of 32 $\times$ 128 if the aspect ratio is less than 4, otherwise, it is resized to $H$ = 32 and $W$ up to 384. Word accuracy is used as the evaluation metric. Data augmentation like rotation, perspective distortion, motion blur and gaussian noise, are randomly performed and the maximum text length is set to 25 during training. The size of the character set $N_c$ is set to 96 for English and 6625 \cite{ppocrv3} for Chinese. The number of attention heads $h$ in sub-string Encoder and Matcher are set to $\frac{C}{32}$ and \textit{2}, respectively.  All models are trained with mixed-precision on 4 RTX 3090 GPUs.

\subsection{Ablation Study}

\noindent\textbf{Effectiveness of Regularization Training (RT)}. As shown in Fig. \ref{fig:regular_training}(b), RT mitigates the sub-string mismatching effectively. With RT, SMTR successfully discriminates two similar sub-strings (''sion.'' and ``plon.'') and recognizes the text accurately. Tab. \ref{tab:regular} quantifies the effectiveness of RT. It takes effects for both short and long text, where a notable 12.07\% accuracy improvement is observed on LTB. The result indicates similar sub-strings are much better identified.

Tab. \ref{tab:regular} also ablates $l_s$ and $\tilde{S}$, the number of regularized sub-strings. The results indicate that $l_s > 5$ is not necessary. In addition, increasing $\tilde{S}$ enhances the accuracy on both short and long text, again validating the positive effect of RT. Therefore, SMTR selects $l_s = 5$ and $\tilde{S} = 2$.

In addition, to qualitatively assess RT in differentiating similar sub-strings, we examine the cosine similarity between $Q_n^{plon.}$, the embedding of sub-string ``sion.'', and others. As shown in the rightmost column in Fig. \ref{fig:regular_training}(a), $Q_n^{plon.}$ has a similarity score of $1.0$ with itself. Without RT, the similarity between $Q_n^{plon.}$ and $Q_n^{sion.}$, the embedding of its similar sub-string ``sion.'', is as high as 0.95, which leads to SMTR not being able to distinguish between them well. In contrast, RT effectively reduces this similarity from 0.95 to 0.61. As shown in the attention map in Fig. \ref{fig:regular_training}(b), this similarity allows SMTR to avoid this distraction and correctly focus on the next character. This example demonstrates that RT successfully pushes away the representations of two similar sub-strings, thereby enhancing SMTR's ability to distinguish between them and improving recognition accuracy.

\noindent\textbf{Effectiveness of Inference Augmentation (IA)}. As shown in Tab. \ref{tab:ltb_result}, IA largely improves the accuracy of SMTR on LTB, with weighted (W-Avg) and arithmetic (A-Avg) average increasing by 3.99\% and 7.17\%, respectively. Notably, accuracy gains from IA become more distinct as the text length increases, where 2.42\%, 6.63\%, and 12.4\% improvements are observed across the three subsets. This trend arises because longer texts are more likely to contain repeating sub-strings. IA effectively eliminates these repetitions, as shown in Tab. \ref{tab:repeatss}, leading to substantial improvements in recognition accuracy.

\begin{table}[t]\footnotesize
\centering
\begin{tabular}{c|c|ccc|c}
\hline
$\tilde{S}$                   & $l_s$   & \textit{Common}  & \textit{U14M}  & \textit{LTB}  & \textit{Avg}   \\
\hline
w/o & 5 & 95.65 & 83.58 & 32.30 & 70.51 \\
1   & 5 & 95.94 & 84.14 & 44.37 & 74.82 \\
\hline
\multirow{5}{*}{2}      & 4 & \textbf{96.04} & 84.51 & 45.24 & 75.26 \\
& 5 & 95.90 & \textbf{85.00} & \textbf{47.00} & \textbf{75.97} \\
& 6 & 95.64 & 84.26 & 45.06 & 74.99 \\
& 7 & 95.77 & 83.80 & 40.55 & 73.37 \\
\hline
\end{tabular}
\caption{Ablation study on regularization training and $l_s$.}
\label{tab:regular}
\end{table}

\begin{table}[t]\footnotesize
\centering
\setlength{\tabcolsep}{4pt}{
\begin{tabular}{c|ccc|cc}
\hline
\multicolumn{1}{c|}{\multirow{2}{*}{Method}} & $L_{[26,35]}$ & $L_{[36,55]}$ & $L_{\geq 56}$ &\multicolumn{1}{c}{\multirow{2}{*}{W-Avg}} &\multicolumn{1}{c}{\multirow{2}{*}{A-Avg}} \\
                  & 3376  & 1147  & 266   &                               &                                 \\
\hline

FocalSVTR               & 51.04 & 25.37 & 0.38  & 42.08                         & 25.59                           \\
AR-STR            & 25.53 & 0.00  & 0.00  & 18.00                         & 8.51                            \\
PARSeq$^\dagger$ \shortcite{BautistaA22PARSeq}           & 0.00  & 0.00  & 0.00  & 0.00                          & 0.00               \\
LISTER$^\dagger$ \shortcite{iccv2023lister}           & 51.16 & 26.59 & 2.26  & 42.56                         & 26.67                           \\
\hline
SMTR w/o IA & 53.06 & 37.05 & 13.16 & 47.00 & 34.42\\
SMTR w/ IA            & \textbf{55.48} & \textbf{43.68} & \textbf{25.56} & \textbf{50.99}                         & \textbf{41.57}                           \\
\hline
\end{tabular}}
\caption{Model comparing results on LTB.}
\label{tab:ltb_result}
\end{table}

\begin{figure}[t]
  \centering
\includegraphics[width=0.47\textwidth]{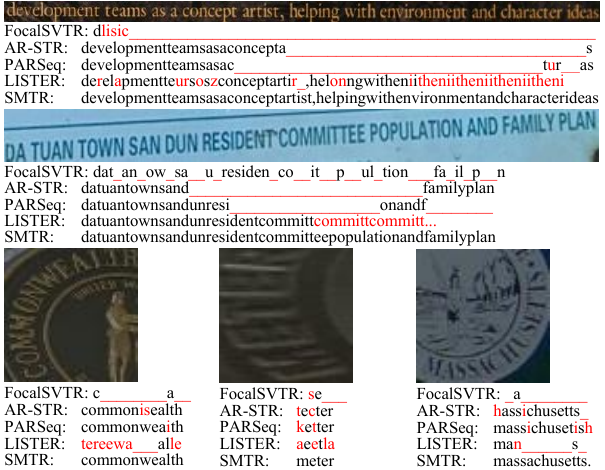} 
\caption{Illustration of the predictions of two long text images from LTB and three short text images from Union14M-Benchmark. In addition, their attention maps are visualized in Supplementary for qualitative analysis.}
\label{fig:goodcase}
\end{figure}

\noindent\textbf{Fair comparison with aligned methods on LTB}. As aforementioned, some STR models, despite being designed for short text recognition, are also capable of recognizing long text, e.g., CTC-based and auto-regressive (AR) decoding ones. We align four of them in Tab. \ref{tab:ltb_result} with SMTR for comparison. Specifically, we uniformly adopt FocalSVTR as the image encoder, keeping their decoders unchanged, and then training them using the same hyper-parameters detailed in section 3.1. The results are presented in Tab. \ref{tab:ltb_result} and Tab. \ref{tab:sota_u14m}, where FocalSVTR is the encoder combining with the naive CTC decoder, AR-STR is the encoder plus with a standard AR decode composed of two transformer units \cite{NIPS2017_attn}, and $^\dagger$ denotes our aligned reproductions.

As seen in Tab. \ref{tab:ltb_result}, SMTR consistently outperforms the compared models in terms of accuracy, and their gaps are sharply enlarged as the text length increases. When inspecting the arithmetic average for $L > 35$, SMTR outperforms FocalSVTR, AR-STR and LISTER by 15.98\%, 33.06\%, and 14.90\%, respectively. The first two long text examples in Fig. \ref{fig:goodcase} show that all four comparing methods suffer from the problem of missing characters. In addition, AR-STR and PARSeq are only able to recognize the front characters due to employing absolute positional encoding and only trained on short text. LISTER encounters the problem of circular recognition due to capturing incorrect neighbor characters. In contrast, SMTR effectively recognizes these instances by sub-string matching. These results convincingly verify the superiority of SMTR in recognizing long text and addressing the OOL challenge.

\begin{table*}[t]\footnotesize
\centering
\setlength{\tabcolsep}{3pt}{
\begin{tabular}{r|ccccccc|cccccccc|c|c}
\hline
\multirow{2}{*}{Method} & \multicolumn{7}{c|}{Common Benchmarks}                                                                & \multicolumn{8}{c|}{Union14M-Benchmark}                                                      &  \multirow{2}{*}{\textit{LTB}}     & \multirow{2}{*}{\begin{tabular}[c]{@{}c@{}}Params\\  ($\times 10^6$)\end{tabular}} \\
& \textit{IC13} & \textit{SVT}  & \textit{IIIT} & \textit{IC15} & \textit{SVTP} & \textit{CUTE} & \textit{Avg} & \textit{CUR} & \textit{MLO} & \textit{ART} & \textit{CTL} & \textit{SAL} & \textit{MLW} & \textit{GEN} & \textit{Avg}         &   &                           \\
\hline

 CRNN \shortcite{shi2017crnn}             & 91.8 & 83.8 & 90.8 & 71.8 & 70.4 & 80.9 & 81.58                                                                       & 19.4  & 4.5                                                       & 34.2     & 44.0        & 16.7    & 35.7                                                   & 60.4    & 30.70                                                                 &   -   & 8.3                                                                                                                 \\
     
SVTR-B* \shortcite{duijcai2022svtr}          & 97.5 & 96.4 & 97.8 & 89.3 & 91.0 & 96.2 & 94.72                                                                       & 85.4  & 87.4                                                      & 68.9     & 79.5        & 84.3    & 79.1                                                   & 81.8    & 80.91                                                                     & - & 24.6                                                                                                                \\
FocalSVTR & 97.3 & 96.3 & 98.2 & 87.4 & 88.4          & 96.2 & 93.97          & 77.7 & 62.4 & 65.7 & 78.6 & 71.6 & 81.3          & 79.2 & 73.80     &  42.1   &  14.7 \\
\hline
DAN \shortcite{wang2020aaai_dan}              & 95.2 & 88.6 & 95.5 & 78.3 & 79.9 & 86.1 & 87.26                                                                       & 46.0  & 22.8                                                      & 49.3     & 61.6        & 44.6    & 61.2                                                   & 67.0    & 50.44                                                               &   0.0     & 27.7                                                                                                                \\
SRN \shortcite{yu2020srn}              & 94.7 & 89.5 & 95.5 & 79.1 & 83.9 & 91.3 & 89.00                                                                       & 49.7  & 20.0                                                      & 50.7     & 61.0        & 43.9    & 51.5                                                   & 62.7    & 48.50                                                               &   0.0     & 54.7                                                                                                                \\
RoScanner \shortcite{yue2020robustscanner}    & 95.7 & 92.4  & 96.8 & 86.4 & 83.9 & 93.8 & 91.50                                                                       & 66.2  & 54.2                                                      & 61.4     & 72.7        & 60.1    & 74.2                                                   & 75.7    & 66.36                                                           &      0.0      & 48.0                                                                                                                  \\
  ABINet \shortcite{fang2021abinet}       & 97.2 & 95.7 & 97.2 & 87.6 & 92.1 & 94.4 & 94.03                                                                       & 75.0  & 61.5                                                      & 65.3     & 71.1        & 72.9    & 59.1                                                   & 79.4    & 69.19                                                               &   0.0    & 36.7  \\
 VisionLAN \shortcite{Wang_2021_visionlan}    & 95.1 & 91.3 & 96.3  & 83.6 & 85.4 & 92.4 & 90.68                                                                       & 70.7  & 57.2                                                      & 56.7     & 63.8        & 67.6    & 47.3                                                   & 74.2    & 62.50                                                             &    0.0      &  32.8                                                                                                                \\

 PARSeq* \shortcite{BautistaA22PARSeq}  & \textbf{98.4} & 98.1 & 98.9 & \textbf{90.1} & 94.3 & \textbf{98.6} & \textbf{96.40} & 87.6 & 88.8 & 76.5 & 83.4 & 84.4 & 84.3 & \textbf{84.9} & 84.26 & 0.0 & 23.8 \\

 MATRN \shortcite{MATRN}       & 97.9 & 96.9 & 98.2 & 88.2 & 94.1 & 97.9 & 95.53                                                                       & 80.5  & 64.7                                                      & 71.1     & 74.8        & 79.4    & 67.6                                                   & 77.9    & 73.71                                                               &    0.0    & 44.2  \\  
 BUSNet* \shortcite{Wei_Zhan_Lu_Tu_Yin_Liu_Pal_2024_busnet}   & 97.8 & 98.1 & 98.3 & 80.2 & 95.3 & 96.5 & 96.06          & 83.0 & 82.3 & 70.8 & 77.9 & 78.8 & 71.2 & 82.6 & 78.10     &  0.0  & 32.0 \\
 OTE \shortcite{Xu_2024_CVPR_OTE}   & 98.0 & 98.0 & 98.1 & 89.1 & \textbf{95.5} & 97.6 & 96.10          & 83.1 & 82.8 & 73.5 & 73.7 & 79.7 & 70.3 & 82.2 & 77.90     &  0.0  & 25.2 \\
\hline

ASTER \shortcite{shi2019aster}            & 92.6 & 88.9 & 94.3 & 77.7 & 80.5 & 86.5 & 86.75                                                                       & 38.4  & 13.0                                                      & 41.8     & 52.9        & 31.9    & 49.8                                                   & 66.7    & 42.07                                                                     & - & 27.2                                                                                                                \\
NRTR \shortcite{Sheng2019nrtr}             & 96.9 & 94.0 & 96.2 & 80.9 & 84.8 & 92.0 & 90.80                                                                       & 49.3  & 40.6                                                      & 54.3     & 69.6        & 42.9    & 75.5                                                   & 75.2    & 58.20                                                                 &   -   & 31.7                                                                                                                \\
SAR \shortcite{li2019sar}              & 96.0 & 92.4 & 96.6 & 82.0 & 85.7 & 92.7 & 90.90                                                                       & 68.9  & 56.9                                                      & 60.6     & 73.3        & 60.1    & 74.6                                                   & 76.0    & 67.20                                                                &  -     & 57.7                                                                                                                \\
 MAERec \shortcite{jiang2023revisiting}        & 97.6 & 96.8 & 98.0  & 87.1 & 93.2 & 97.9 & 95.10                                                                       & 81.4  & 71.4                                                      & 72.0     & 82.0        & 78.5    & 82.4                                                   & 82.5    & 78.60                                                           &    -        & 35.7                                                                                                                \\
 CDistNet* \shortcite{zheng2024cdistnet} &97.8 	&97.1 	&98.7 	&89.6 	&93.5 	&96.9 	&95.59 	&81.7 	&77.1 	&72.6 	&78.2 	&79.9 	&79.7 	&81.1 	&78.62 & - & 65.5\\

 LISTER* \shortcite{iccv2023lister}   & 97.4 & 98.1 & 98.2 & 89.2 & 93.5 & 95.5 & 95.33          & 71.6 & 55.9 & 68.9 & 76.4 & 68.1 & 80.2 & 80.9 & 71.72     & 42.6    & 49.9 \\
  AR-STR    & 98.1 & 98.3 & \textbf{99.0} & 89.8 & 94.2          & 97.9 & 96.22          & 88.1 & \textbf{91.2} & 76.2 & 83.3 & 83.2 & 85.6          & \textbf{84.9} & 84.62     &  18.0   & 19.3 \\

\hline
SMTR    & 98.3 & 97.4 & \textbf{99.0} & \textbf{90.1} & 92.7          & 97.9 & 95.90          & \textbf{89.1} & 87.7 & \textbf{76.8} & \textbf{83.9} & \textbf{84.6} & \textbf{89.3}          & 83.7 & \textbf{85.00}       &  \textbf{51.0} &      15.8       \\
\hline
\end{tabular}}
\caption{Results on short text benchmarks and LTB when the models are trained on Union14M-L \shortcite{jiang2023revisiting} datasets. ''-'' means the model cannot recognize long text directly. For short text benchmarks, we use the next inference. * denotes that the results on short benchmarks  is obtained by training the model on Union14M-L using the code they released.}
\label{tab:sota_u14m}
\end{table*}

\subsection{Comparison with State-of-the-arts}

\noindent\textbf{Results on English Benchmarks}. We apply SMTR to Common benchmarks, Union14M-Benchmark and LTB, and compare it with existing STR models. The results are in Tab. \ref{tab:sota_u14m}. Compared with PARSeq, one leading method in short text benchmarks, SMTR exhibits quite similar accuracy. In addition, it correctly recognizes half of LTB text, which is incapable for PARSeq. Note that AR-STR also achieves very competitive results on short text benchmarks. However, its weak extrapolation capability restricts the performance on LTB. On the other hand, FocalSVTR performs decently on short text benchmarks, but on LTB it performs on par with LISTER, which is a dedicated model for long text recognition. The results above are in line with the observation in Fig. \ref{fig:fig1}. To sum, by introducing a novel sub-string matching-based recognition paradigm, SMTR well recognizes both short and long text, which would be a promising property for many STR-related applications.

In the bottom of Fig. \ref{fig:goodcase} we also present the recognition of three challenging short text images. CTC-based FocalSVTR and LISTER show severe mis-recognition due to their reliance on a priori of regular text. In contrast, AR-STR and PARSeq can recognize a majority of the text, encountering errors only when dealing with particularly challenging characters. SMTR accurately identifies all the characters, showing robustness in handling challenging short text.

\noindent\textbf{Results on Chinese Benchmarks}. The OOL challenge is pervasive across languages. To verify the multilingual adaptability of SMTR, we conduct evaluations on the Chinese text recognition benchmark (CTR) \cite{chen2021benchmarking}. In line with CCR-CLIP \cite{yuICCV2023clipctr}, SMTR refrains from utilizing data augmentation (w/o Aug) to ensure a fair comparison. As presented in Tab. \ref{tab:ch_all}, SMTR outperforms LISTER by 6.83\% on \textit{Scene} subset. Note that SMTR gets this result without training on long text while LISTER has no imposed length limitation during training. Despite this discrepancy, SMTR still maintains an accuracy gain of 10.61\% over LISTER in Chinese long text ($L_{> 25}$). Meanwhile, SMTR achieves a new state-of-the-art on CTR, boosting an accuracy gain of 3.53\% compared to CAM \cite{yang2024class_cam}, the previous best method. These results demonstrate SMTR's great adaptability in multilingual recognition.

\begin{table}[t]\footnotesize
\centering
\setlength{\tabcolsep}{2.3pt}{
\begin{tabular}{r|cccc|c|c|c}
\hline
Method        & \textit{Scene}         & \textit{Web}           & \textit{Doc} & \textit{HW}  & Avg  & $L_{> 25}$          & \begin{tabular}[c]{@{}c@{}}Params\\ $\times 10^6$\end{tabular} \\
\hline
CRNN \shortcite{shi2017crnn}         & 53.4          & 57.0            & 96.6          & 50.8          & 64.45    &   -   & 12.4   \\
ASTER \shortcite{shi2019aster}        & 61.3          & 51.7          & 96.2          & 37.0            & 61.55     &  -   & 27.2   \\
MORAN \shortcite{pr2019MORAN}        & 54.6          & 31.5          & 86.1          & 16.2          & 47.10     &   -   & 28.5   \\
SAR \shortcite{li2019sar}          & 59.7          & 58.0            & 95.7          & 36.5          & 62.48    &   -   & 27.8   \\
SEED \shortcite{cvpr2020seed}         & 44.7          & 28.1          & 91.4          & 21.0            & 46.30   &   -     & 36.1   \\
MASTER \shortcite{pr2021MASTER}       & 62.8          & 52.1          & 84.4          & 26.9          & 56.55     &  -   & 62.8   \\
ABINet \shortcite{fang2021abinet}       & 66.6          & 63.2          & 98.2          & 53.1          & 70.28    &  0.0    & 53.1   \\
TransOCR \shortcite{cvpr2021TransOCR}     & 71.3          & 64.8          & 97.1          & 53.0            & 71.55   &   -    & 83.9   \\
SVTR-B \shortcite{duijcai2022svtr}       & 71.7          & 73.8          & 98.2          & 52.2          & 73.98     &   -  & 26.3   \\
CCR-CLIP \shortcite{yuICCV2023clipctr}     & 71.3          & 69.2          & 98.3          & 60.3          & 74.78   &    -   & 62.0     \\
LISTER \shortcite{iccv2023lister} & 73.0 & - & - & - & - & 35.00 &55.0 \\
DCTC \shortcite{Zhang_Lu_Liao_Huang_Li_Wang_Peng_2024_DCTC}     & 73.9          & 68.5          & \textbf{99.4}          & 51.0          & 73.20       & -  & 40.8     \\
CAM \shortcite{yang2024class_cam}     & 76.0          & 69.3          & 98.1          & 59.2          & 76.80       & -  & 135     \\
\hline
SMTR w/o Aug & 79.8 & 80.6 & 99.1 & 61.9 & 80.33 & 45.61 & 20.8   \\
SMTR w/ Aug         & \textbf{83.4}          &   \textbf{83.0}            &  99.3             &    \textbf{65.1}           &  \textbf{82.68}       &  \textbf{49.41}     &   20.8    \\
\hline
\end{tabular}}
\caption{Results of different models on CTR dataset. $L_{> 25}$ means long text recognition results on \textit{Scene}.}
\label{tab:ch_all}
\end{table}

\section{Conclusion}

In this paper, we point out that, in real applications, STR models sometimes have to recognize long text images and existing models trained on short text images cannot accomplish this task well. We term this the OOL text recognition challenge. We construct the LTB dataset for long text recognition assessment, and propose a novel SMTR that employs a sub-string-matching-based paradigm to overcome this challenge. SMTR implements the recognition by iteratively predicting the next and previous characters of a sub-string. It is capable of recognizing both short and long text. To make the recognition effective in this new paradigm, we introduce regularization training to suppress distractions caused by similar sub-strings, and inference augmentation to alleviate confusion caused by repeated sub-strings and improve recognition efficiency. Experimental results show that SMTR not only surpasses existing methods by a clear margin on LTB, but is also highly competitive on challenging short text benchmarks. Nevertheless, SMTR employs an iterative recognition paradigm in inference thus the speed is not fast. Hence, our future research will be devoted to addressing this problem.  

\section*{Acknowledgments}

This work was supported by National Key R\&D Program of China (No. 2022YFB3104703) and in part by the National Natural Science Foundation of China (No. 32341012).

\bibliography{aaai25}

\appendix
\clearpage

\setcounter{page}{1}

\begin{table}[t]\footnotesize
\centering
\setlength{\tabcolsep}{3pt}{
\begin{tabular}{c|cc|c|cc}
\hline
Source    & Train & Test & Source         & Train & Test \\
\hline
RCTW \shortcite{shi2017icdar2017rctw}      & 173           & 1         & MTWI \shortcite{he2018icpr2018mtwi}           & 516           & 77        \\
LSVT \shortcite{sun2019icdarlsvt}      & 968           & 111       & UberText \shortcite{zhang2017ubertext}           & 465           & 40        \\
ArT \shortcite{chng2019icdar2019art}       & 28            & 4         & COCOText \shortcite{veit2016cocotext}     & 23            & 0         \\
IntelOCR \shortcite{krylov2021openintelocr}  & 792           & 111       & ReCTS \shortcite{zhang2019icdarrects}         & 3             & 1         \\
TextOCR \shortcite{singh2021textocr}  & 140           & 16        & KAIST  \shortcite{jung2011touchkaist}        & 11            & 0         \\
HierText \shortcite{long2022towardshiertext} & 572           & 82        & NEOCR  \shortcite{nagy2012neocr}        & 59            & 10        \\
MLT19 \shortcite{nayef2019icdar2019mlt19}    & 25            & 1         & CTW1500  \shortcite{yuliang2017detectingctw1500}      & 551           & 0         \\
IIIT-ILST \shortcite{mathew2017benchmarkingiiitilst} & 1             & 0         & \textbf{Total} & \multicolumn{2}{c}{4,789} \\
\hline
\end{tabular}}
\caption{Composition details of LTB.}
\label{tab:ltb}
\end{table}

\begin{figure*}[h]
  \centering
\includegraphics[width=0.88\textwidth]{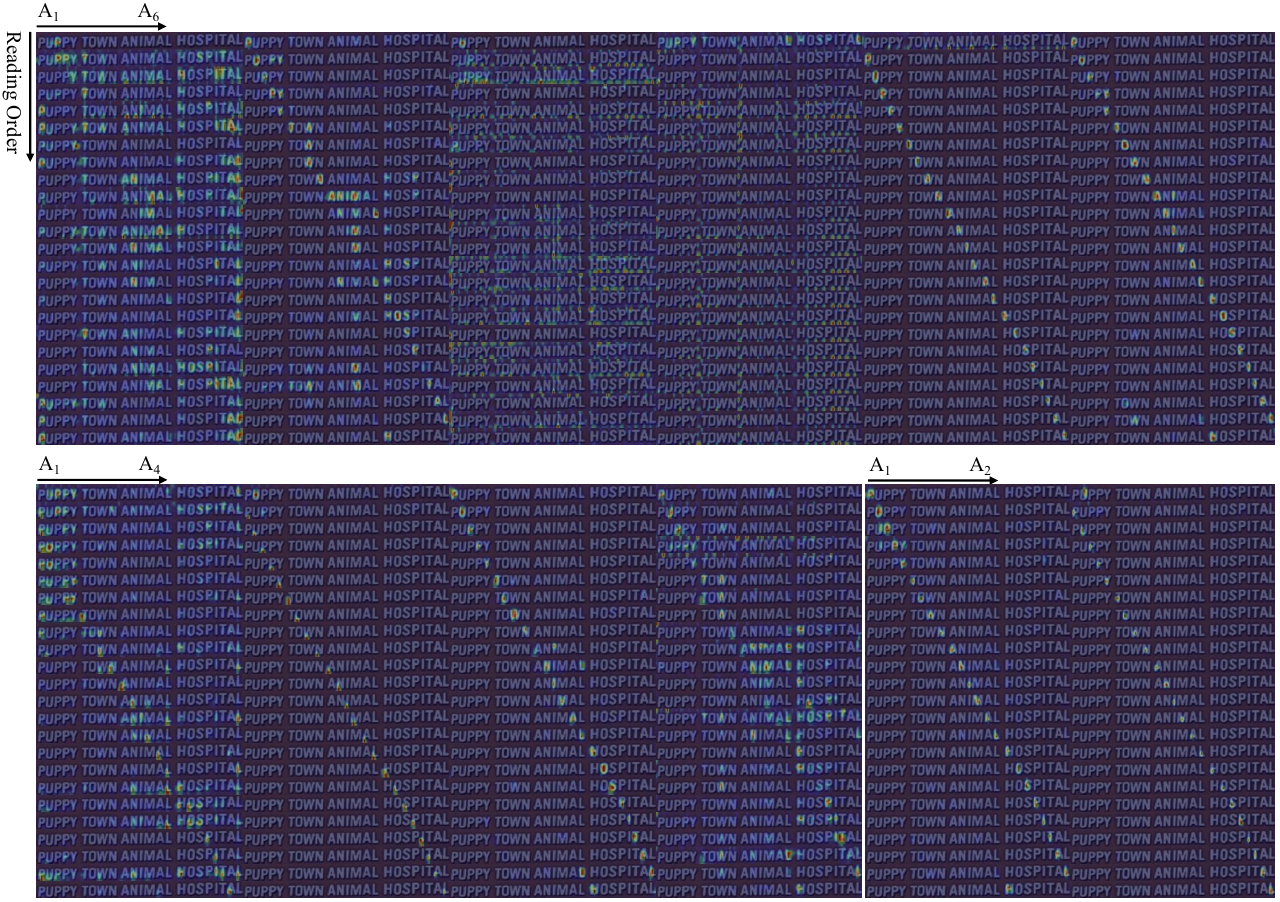} 
\caption{The attention maps $A_i$ in sub-string matcher when $h$ is set to 6, 4 and 2.}
\label{fig:h642}
\end{figure*}

\begin{algorithm}[t] \footnotesize
\caption{Inference Augmentation Pseudo-code in Python}
\PyCode{def InferenceLongText(Img):} \\
\Indp
    \PyComment{Img: Input Image, $3 \times H \times W$.} \\
    \PyCode{Img1,Img2,Img3 = Slice(Img)} \\
    \PyCode{R1 = Inference(Img1,$T_n$)} \\
    \PyComment{[::-1] for inverted a list.} \\
    \PyCode{R2 = Inference(Img2,$T_p$)[::-1]} \\
    \PyCode{$T_{Next}$ = R1[:-($l_s$+1)]} \\
    \PyCode{$T_{Pre}$ = R2[$l_s$+1:]} \\
    \PyCode{Ss = R1[-($l_s$+1):-1]} \\
    \PyCode{Se = R2[1:$l_s$+1]} \\
    \PyCode{R3 = Inference(Img3,$T_n$,Ss,Se)} \\
    \PyCode{$T_{Mid}$ =R3[:-($l_s$-1)]} \\
    \PyCode{Result = $T_{Next}$+Ss+$T_{Mid}$+Se+$T_{Pre}$}\\
    \PyCode{return Result} \\
    \PyComment{Result is the recognition.} \\
\Indm
\label{alg:inference_long}
\end{algorithm}

\begin{figure}[t]
  \centering
\includegraphics[width=0.44\textwidth]{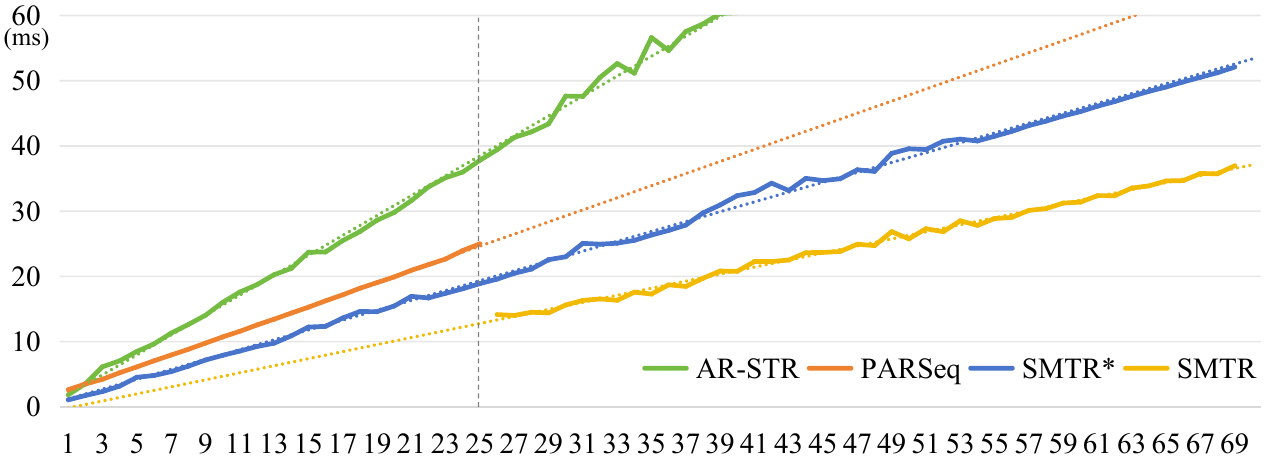} 
\caption{Inference speeds of different methods. SMTR$^*$ means not using inference augmentation. For a fair comparison, the same encoder is employed for all the methods and only time consumption of the decoding stage is considered.}
\label{fig:speed}
\end{figure}

\section{Pseudo-code of Inference Augmentation}

Algorithm \ref{alg:inference_long} presents the implementation of inference augmentation (IA) in Python style. Specifically, the \texttt{Slice} operation splits an image into three sub-images as shown in Fig. 4. It is then followed by bi-directional inference using \texttt{Inference} to obtain the recognition results of \texttt{Img1} and \texttt{Img2}. In the following, the third \texttt{Inference} is performed on \texttt{Img3} with sub-strings from both sides given, which can correct mis-recognition caused by \texttt{Slice}. Finally, the full long text recognition result is obtained with a straightforward post-processing. The operation \texttt{Inference} is detailed in Algorithm \ref{alg:inference}.

By leveraging IA, SMTR not only achieves enhanced recognition accuracy for long texts but also significantly accelerates inference speed. As shown in Fig. \ref{fig:speed}, IA improves SMTR's efficiency in long text recognition. This improvement is primarily attributed to IA's split-merge processing mechanism, which substantially reduces the number of iterations required for recognizing lengthy texts.

\section{Details of the Sub-String Matcher Module}

We ablate the number of heads ($h$), and the role of residual and MLP operations in the multi-head attention described in Sec. 3.3. As shown in Tab. \ref{tab:network}, residual and MLP exhibit effectiveness in short text recognition no matter whether the head number ($h$) is set to either 2 or 12. However, inverse results are observed in long text recognition. We argue that the possible explanation is as follows. When the residual is applied, $\tilde{Y_n}$ is adapted as $\text{Classifier}(\text{Matcher}(Q_{n}, E_I)+Q_n)$. This suggests that the content of a sub-string is directly involved in character features and facilitates recognition by exploiting fixed sub-string patterns during short text training. For instance, the next character of sub-string ``Cente'' is more likely to be ``r''. Previously, these fixed patterns are considered as extra linguistic information. Nevertheless, in long texts, sub-string patterns become more diverse, and the persistence of fixed sub-string patterns derived from short text training may be detrimental to long text recognition. Therefore, removing the residual and MLP enhances the recognition of long text.

Another interesting observation is the choice of $h$. When setting $h$ to 2, there is an improvement of 2.26\% in accuracy on LTB compared to setting $h$ to 12. Through the visualization of attention maps at various $h$ (Fig. \ref{fig:h642}), it is observed that only two heads are actually activated when $h$ exceeds 2, which in part explains why employing two heads gets the best results. In addition, despite getting slightly worse results on short text benchmarks when setting $h$ to 2 and getting rid of both residual and MLP, this setting still gets the best overall performance across the three benchmarks. Therefore our SMTR chooses the setting above. These ablations also indicate that only a simple and lightweight design is required for the sub-string matcher module. 

\begin{table}[t]\footnotesize
\centering
\setlength{\tabcolsep}{4.5pt}{
\begin{tabular}{c|c|c|ccc|c}
\hline
           $h$          & Residual           & MLP                & Common  & U14M  & LTB  & Avg   \\
\hline
\multirow{2}{*}{12} & 1                  & 1                  & \textbf{96.26} & 85.33 & 43.33 & 74.97 \\
                     & 1                  & 0                  & 96.07 & 84.65 & 44.54 & 75.09 \\
\hline
12                  & \multirow{7}{*}{0} & \multirow{7}{*}{0} & 95.80 & 84.52 & 44.74 & 75.02 \\
8                   &                    &                    & 96.02 & 85.03 & 45.60 & 75.55 \\
6                   &                    &                    & 95.81 & 84.47 & 45.24 & 75.17 \\
4                   &                    &                    & 95.79 & 84.40 & 46.39 & 75.53 \\
3                   &                    &                    & 95.78 & 84.18 & 44.08 & 74.68 \\
2                   &                    &                    & 95.90 & 85.00 & \textbf{47.00} & \textbf{75.97} \\
1                   &                    &                    & 95.68 & 84.58 & 46.06 & 75.44 \\
\hline
\multirow{2}{*}{2}  & 1                  & 0                  & 96.03 & 84.88 & 46.58 & 75.83 \\
                     & 1                  & 1                  & 96.04 & \textbf{85.41} & 45.18 & 75.54 \\
\hline
\end{tabular}}
\caption{Ablation study on the number of $h$, the role of residual and MLP in the sub-string matcher module.}
\label{tab:network}
\end{table}

\begin{table*}[!h]\footnotesize
\centering
\setlength{\tabcolsep}{2.8pt}{
\begin{tabular}{r|ccccccc|cccccccc|c|c}
\hline
\multirow{2}{*}{Method} & \multicolumn{7}{c|}{Common Benchmarks}                                                                & \multicolumn{8}{c|}{Union14M-Benchmark}                                                      &  \multirow{2}{*}{\textit{LTB}}     & \multirow{2}{*}{\begin{tabular}[c]{@{}c@{}}Params\\  ($\times 10^6$)\end{tabular}} \\
& \textit{IC13} & \textit{SVT}  & \textit{IIIT} & \textit{IC15} & \textit{SVTP} & \textit{CUTE} & \textit{Avg} & \textit{CUR} & \textit{MLO} & \textit{ART} & \textit{CTL} & \textit{SAL} & \textit{MLW} & \textit{GEN} & \textit{Avg}         &   &                           \\
\hline

 CRNN \shortcite{shi2017crnn}             & 91.1 & 81.6 & 82.9 & 69.4 & 70.0 & 65.5 & 76.75                                                                       & 7.5   & 0.9                                                       & 20.7     & 25.6        & 13.9    & 25.6                                                   & 32.0    & 18.03                                                                  & -    & 8.3                                                                                                                 \\
 SVTR-B* \shortcite{duijcai2022svtr}          & 97.1 & 91.5 & 96.0 & 85.2 & 89.9 & 91.7 & 91.90                                                                       & 69.8  & \textbf{37.7}                                                      & 47.9     & 61.4        & 66.8    & 44.8                                                   & 61.0    & 55.63                                                                   &  -  & 24.6                                                                                                                \\
 DCTC \shortcite{Zhang_Lu_Liao_Huang_Li_Wang_Peng_2024_DCTC} & 97.4          & 93.7         & 96.9 & 87.3         & 88.5          & 92.3          & 92.68                      & -                 & -                                                      & -                 & -                 & -                 & -                                                   & -                 & -                                              &  -  & 40.8      \\
\hline
 RoScanner \shortcite{yue2020robustscanner}   & 94.8 & 88.1 & 95.3 & 77.1 & 79.5 & 90.3 & 87.52                                                                       & 43.6  & 7.9                                                       & 41.2     & 42.6        & 44.9    & 46.9                                                   & 39.5    & 38.09                                                                  &  0.0   & 48.0                                                                                                                  \\
 SRN \shortcite{yu2020srn}             & 95.5 & 91.5 & 94.8 & 82.7 & 85.1 & 87.8 & 89.57                                                                       & 63.4  & 25.3                                                      & 34.1     & 28.7        & 56.5    & 26.7                                                   & 46.3    & 40.14                                                                   & 0.0   & 54.7                                                                                                                \\
 VisionLAN \shortcite{Wang_2021_visionlan}       & 95.7 & 91.7 & 95.8 & 83.7 & 86.0 & 88.5 & 90.23                                                                       & 57.7  & 14.2                                                      & 47.8     & 48.0        & 64.0    & 47.9                                                   & 52.1    & 47.39                                                           &    0.0        & 32.8                                                                                                                \\
ABINet \shortcite{fang2021abinet}          & 97.4 & 93.5 & 96.2 & 86.0 & 89.3 & 89.2 & 91.93                                                                       & 59.5  & 12.7                                                      & 43.3     & 38.3        & 62.0    & 50.8                                                   & 55.6    & 46.03                                                                &   0.0    & 36.7                                                                                                                \\

 LPV-B* \shortcite{ijcai2023LPV}           & 97.6 & 94.6 & 97.3 & 87.5 & 90.9 & 94.8 & 93.78                                                                       & 68.3  & 21.0                                                      & \textbf{59.6}     & 65.1        & 76.2    & 63.6                                                   & 62.0    & 59.40                                                                  & 0.0    & 35.1                                                                                                                \\
 MATRN  \shortcite{MATRN}       & 97.9          & 95.0         & 96.6 & 86.6         & 90.6          & 93.5          & 93.37                      & 63.1                 & 13.4                                                      & 43.8                 & 41.9                 & 66.4                 & 53.2                                                   & 57.0                 & 48.40                                              &  0.0  & 44.2      \\
 PARSeq* \shortcite{BautistaA22PARSeq}          & 97.0 & 93.6 & 97.0 & 86.5 & 88.9 & 92.2 & 92.53                                                                       & 63.9  & 16.7                                                      & 52.5     & 54.3        & 68.2    & 55.9                                                   & 56.9    & 52.62                                                                &  0.0     & 23.8                                                                                                                \\
 MGP-STR*  \shortcite{mgpstr}     & 97.3          & 94.7         & 96.4 & 87.2         & 91.0          & 90.3          & 92.82                                                                     &55.2 &14.0& 52.8&48.5&65.2&48.8& 59.1& 49.09 &  0.0 & 148       \\
   CAM* \shortcite{yang2024class_cam} & 97.2          & \textbf{96.1}         & \textbf{97.4} & 87.8         & 90.6          & 92.4          & 93.58                      & 63.1                 & 19.4                                                      & 55.4                 & 58.5                 & 72.7                 & 51.4                                                   & 57.4                 & 53.99                                              &  0.0  & 135      \\
BUSNet \shortcite{Wei_Zhan_Lu_Tu_Yin_Liu_Pal_2024_busnet} & \textbf{98.3}          & 95.5         & 96.2 & 87.2         & \textbf{91.8}          & 91.3          & 93.38                      & -                 & -                                                      & -                 & -                 & -                 & -                                                   & -                 & -                                              &  0.0  & 56.8      \\
 OTE \shortcite{Xu_2024_CVPR_OTE} & 97.4          & 95.5         & 96.4 & 87.2         & 89.6          & 92.4          & 93.08                      & -                 & -                                                      & -                 & -                 & -                 & -                                                   & -                 & -                                              &  0.0  & 25.2      \\
\hline
 ASTER \shortcite{shi2019aster}           &   90.8   & 90.0 & 93.3 & 74.7 & 80.2 & 80.9 &  84.98                                                                           & 34.0  & 10.2                                                      & 27.7     & 33.0        & 48.2    & 27.6                                                   & 39.8    & 31.50                                                               &  -      & 27.2                                                                                                                \\
 NRTR \shortcite{Sheng2019nrtr}            & 95.8 & 91.5 & 90.1 & 79.4 & 86.6 & 80.9 & 87.38                                                                       & 31.7  & 4.4                                                       & 36.6     & 37.3        & 30.6    & 54.9                                                   & 48.0    & 34.79                                                                   &  -  & 31.7                                                                                                                \\
 SAR \shortcite{li2019sar}             & 91.0 & 84.5 & 91.5 & 69.2 & 76.4 & 83.5 & 82.68                                                                       & 44.3  & 7.7                                                       & 42.6     & 44.2        & 44.0    & 51.2                                                   & 50.5    & 40.64                                                                    & -  & 57.7                                                                                                                \\

CornerTrans*  \shortcite{xie2022toward}   & 97.8          & 94.6          & 95.9 & 86.5          & 91.5          & 92.0          & 93.05                                                                  &62.9&18.6&56.1&58.5&68.6&59.7&61.0&   55.07 & -  & 86.0                       \\
LISTER* \shortcite{iccv2023lister} & 97.9 & 93.8 & 96.9 & 87.5 & 89.6 & 90.6 & 92.72 & 56.5 & 17.2 & 52.8 & 63.5 & 63.2 & 59.6 & 65.4 & 54.05 & 24.1 & 49.9 \\
CDistNet* \shortcite{zheng2024cdistnet}         & 97.4 & 93.5 & 96.4 & 86.0 & 88.7 & 93.4 & 92.57                                                                       & 69.3  & 24.4                                                      & 49.8     & 55.6        & 72.8    & 64.3                                                   & 58.5    & 56.38                                                                 & -      & 65.5         \\

\hline
 SMTR & 97.4 & 94.9 & \textbf{97.4} & \textbf{88.4} & 89.9 & \textbf{96.2} & \textbf{94.02} & \textbf{74.2} & 30.6 & 58.5 & \textbf{67.6} & \textbf{79.6} & \textbf{75.1} & \textbf{67.9} & \textbf{64.79} & \textbf{39.6} &  15.8 \\
\hline
\end{tabular}}
\caption{Results of SMTR and existing models on short text benchmarks and LTB when trained on synthetic datasets. * represents that the result is evaluated on Union14M-Benchmark using the model they released.}
\label{tab:sota_syn}
\end{table*}

\section{Results when trained on synthetic datasets}

We also conduct experiments by training SMTR and other STR models on synthetic datasets \cite{Synthetic,jaderberg14synthetic} and evaluate them using real-world benchmarks, which is also a typical evaluation protocol and can fully assess the effectiveness of SMTR. The results are presented in Tab. \ref{tab:sota_syn}. The observations are basically in accordance with those in Tab. \ref{tab:sota_u14m}. SMTR surpasses the previous best model by 0.54\% and 4.69\% on the two short text benchmarks, respectively. Moreover, SMTR advances LISTER remarkably on LTB. The results again demonstrate the effectiveness of SMTR.

\section{Visualizations of attention maps}

\begin{figure}[h]
  \centering
\includegraphics[width=0.48\textwidth]{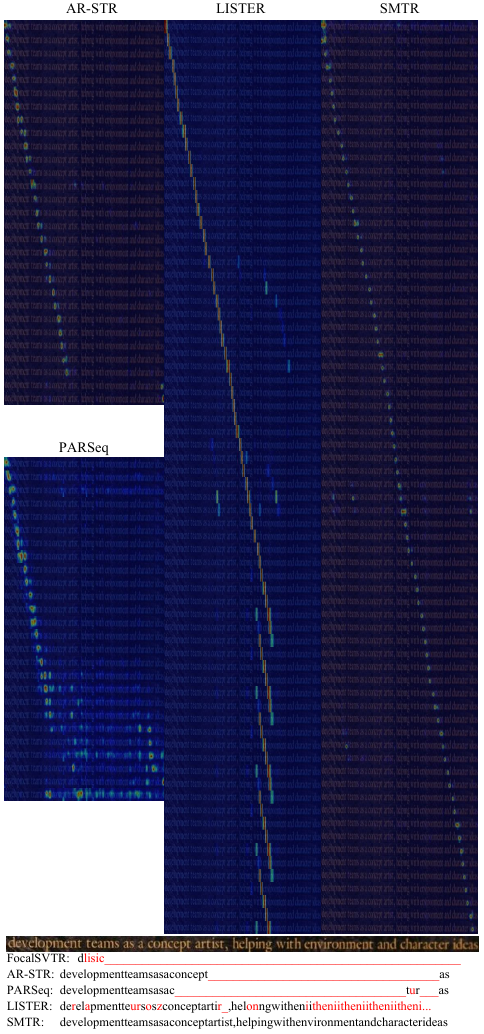} 
\caption{Attention map visualizations of different models and their predictions on the first long text instance.}
\label{fig:long1}
\end{figure}

\begin{figure}[h]
  \centering
\includegraphics[width=0.48\textwidth]{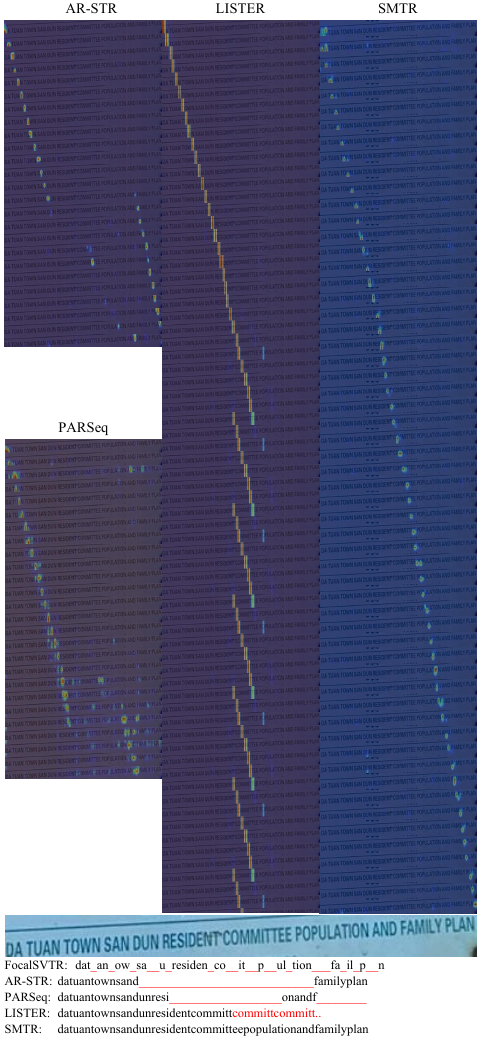} 
\caption{Attention map visualizations of different models and their predictions on the second long text instance.}
\label{fig:long2}
\end{figure}

\begin{figure*}[h]
  \centering
\includegraphics[width=0.88\textwidth]{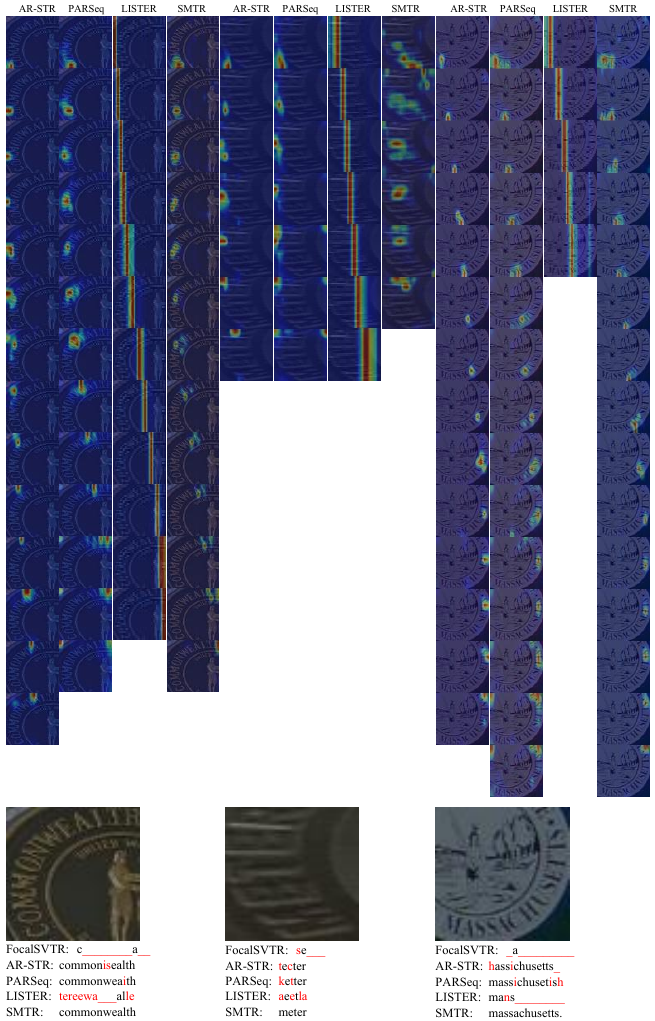} 
\caption{Attention map visualizations of different models and their predictions on the three short text instances.}
\label{fig:shorttext}
\end{figure*}

Fig. \ref{fig:goodcase} displays the recognition results of two long texts and three short texts. We visualize attention maps of these five samples in Fig. \ref{fig:long1}, Fig. \ref{fig:long2} and Fig. \ref{fig:shorttext}. These visualizations vividly explain how the text is recognized for four of the compared models. As shown in Fig. \ref{fig:long1} and \ref{fig:long2}, the attention maps of sinAR and PARSeq, when recognizing long texts, tend to skip some characters, resulting in character missing. On the other hand, LISETR's attention maps show a circular focusing phenomenon, which is in line with its repeated circular recognition errors. Furthermore, LISETR implicitly assumes the text is horizontally displayed, facing substantial recognition challenges for curved or rotated samples, as depicted in Fig. \ref{fig:shorttext}. In contrast, SMTR leverages the proposed sub-string matching paradigm. It precisely focuses on character positions according to the text reading order, and achieves accurate recognition in both long and short text. This advantage underscores SMTR's superior performance over other attention-based methods in addressing OOL challenges and recognizing text of arbitrary length.

\section{Bad cases of SMTR}

\begin{figure*}[t]
  \centering
\includegraphics[width=0.98\textwidth]{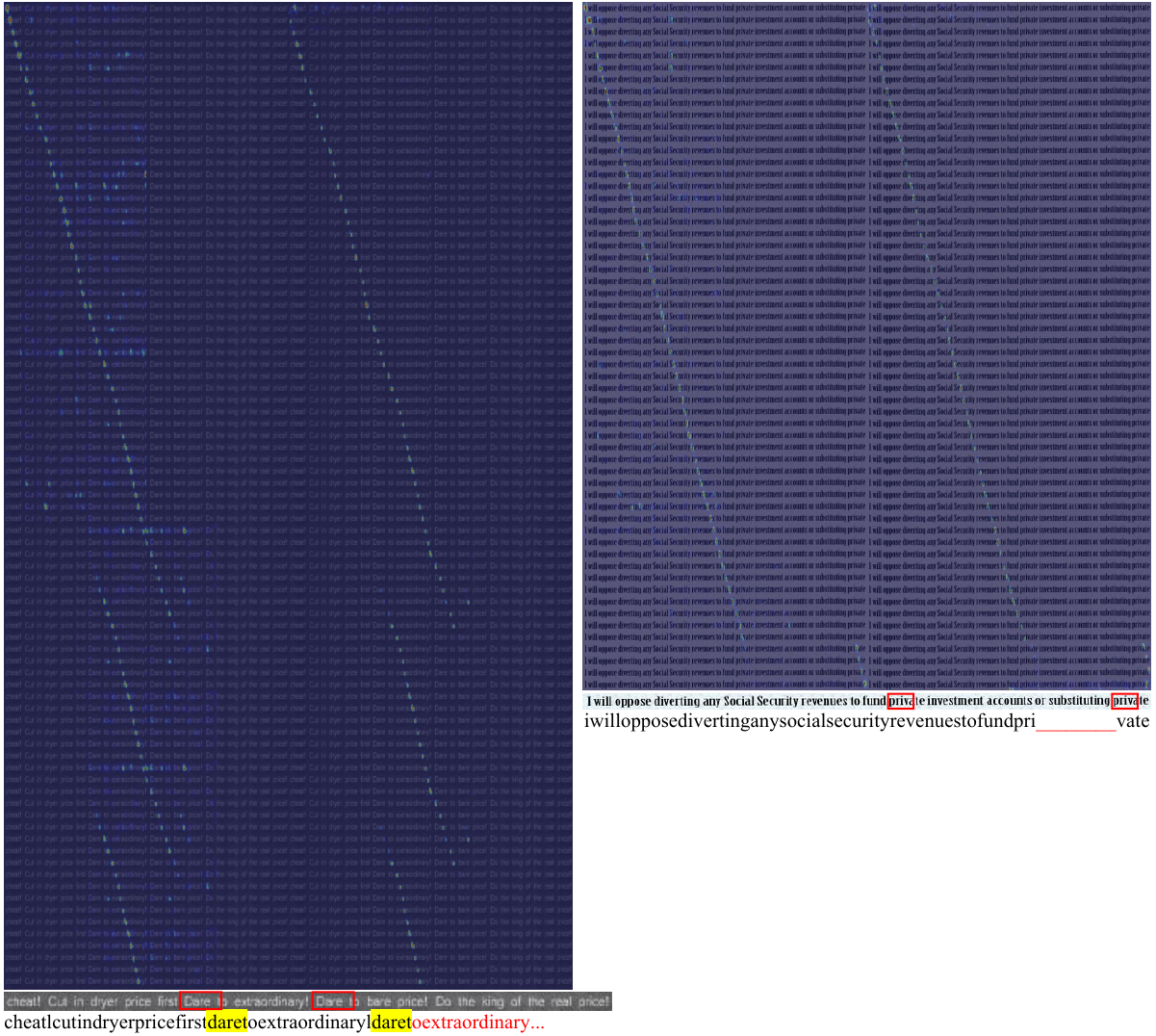} 
\caption{Two bad cases of SMTR. The presence of repeated sub-strings in the text is responsible for the recognition error.}
\label{fig:badcase}
\end{figure*}

There still are a few recognition errors for SMTR. By analyzing the prediction results, we find that the errors are mainly caused by repeated sub-strings. For example, the two instances illustrated in Fig. \ref{fig:badcase}. Although the inference augmentation significantly reduces the percentage of repeated sub-strings, there still are a few exceptions. How to address these instances is also a topic worthy of further study.

\end{document}